\documentclass[journal]{IEEEtran}

\usepackage{algorithm}
\usepackage{algorithmic}
\usepackage{amsmath}
\usepackage{amssymb}
\usepackage{graphicx}
\usepackage{color}
\usepackage{subfigure}
\usepackage{flushend}

\DeclareMathOperator*{\argmin}{argmin}
\DeclareMathOperator*{\argmax}{argmax}

\newcommand{\Lagr}{\mathcal{L}}
\newlength\myindent
\begin{document}
\title{Right for the Right Reason: Training Agnostic Networks}
\author{Sen Jia, Thomas Lansdall-Welfare, and Nello Cristianini\\
Intelligent Systems Laboratory, University of Bristol, Bristol BS8 1UB, UK\\
\{sen.jia, thomas.lansdall-welfare, nello.cristianini\}@bris.ac.uk}

\maketitle              % typeset the title of the contribution

\begin{abstract}
We consider the problem of a neural network being requested to classify images (or other inputs) without making implicit use of a ``protected concept'', that is a concept that should not play any role in the decision of the network. Typically these concepts include information such as gender or race, or other contextual information such as image backgrounds that might be implicitly reflected in unknown correlations with other variables, making it insufficient to simply remove them from the input features. {\color{black}In other words, making accurate predictions is not good enough if those predictions rely on information that should not be used: predictive performance is not the only important metric for learning systems.}
We apply a method developed in the context of domain adaptation to address this problem of ``being right for the right reason'', where we request a classifier to make a decision in a way that is entirely `agnostic' to a given protected concept ({\em e.g.} gender, race, background etc.), even if this could be implicitly reflected in other attributes via unknown correlations. After defining the concept of an `agnostic model', we demonstrate how the Domain-Adversarial Neural Network can remove unwanted information from a model using a gradient reversal layer.
\end{abstract}

\section{Introduction}
Data-driven Artificial Intelligence (AI) is behind the new generation of success stories in the field, and is predicated not just on a few technological breakthroughs, but on a cultural shift amongst its practitioners: namely the belief that predictions are more important than explanations, and that correlations count more than causations {\color{black} \cite{cristianini2014current,halevy2009unreasonable}}. Powerful black-box algorithms have been developed to sift through data and detect any possible correlation between inputs and intended outputs, exploiting anything that can increase predictive performance. Computer vision (CV) is one of the fields that has benefited the most from this choice, and therefore can serve as a test bed for more general ideas in AI.

{\color{black}This paper targets the important problem of ensuring trust in AI systems.} Consider a case as simple as object classification. It is true that exploiting contextual clues can be beneficial in CV and generally in AI tasks. After all, if an algorithm thinks it is seeing an elephant (the object) in a telephone box (the context), or Mickey Mouse driving a Ferrari, it is probably wrong. This illustrates that even though your classifier might have an opinion about the objects in an image, the context around it can be used to improve your performance ({\em e.g.} telling you that it is unlikely to be an elephant inside a telephone box), as shown in many recent works \cite{chu2018deep,li2017attentive,redmon2016you}.

However making predictions based on context can also lead to problems and creates various concerns, one of which is the use of classifiers in ``out of domain'' situations, a problem that leads to research questions in domain adaptation \cite{ganin2016domain,wulfmeier2017addressing}. Other concerns are also created around issues of bias, {\em e.g.} classifiers incorporating biases that are present in the data and are not intended to be used \cite{caliskan2017semantics}, which run the risk of reinforcing or amplifying cultural (and other) biases \cite{zhao2017men}. Therefore, both predictive accuracy and fairness are heavily influenced by the choices made when developing black-box machine-learning models.

%Ideally, we would like our AI models to be `right for the right reasons' \cite{rftrr}, which is a shift from the aforementioned belief that made the recent AI revolution possible, and  clashes with the current set of practices that we use to source our training data.

Since the limiting factor in training models is often sourcing labelled data, a common  choice is to resort to {\color{black} reusing existing data for a new purpose, such as using web queries to generate training data, and employing various strategies to annotate labels, {\em i.e.}} using proxy signals that are expected to be somewhat correlated to the intended target concept \cite{deng2009imagenet,jia2016gender}. These methods come with no guarantees of being unbiased, or even to reflect the deployment conditions necessarily, with any data collected ``in the wild'' \cite{halevy2009unreasonable,huang2007labeled} carrying with it the biases that come from the wild.

To address these issues, a shift in thinking is needed, from the aforementioned belief that predictions are more important than explanations, to ideally developing models that make predictions that are right for the right reason{\color{black}, and consider other metrics, such as fairness, transparency and trustworthiness, as equally important as predictive performance}. This means that we want to ensure that certain protected concepts are not used as part of making critical decisions ({\em e.g.} decisions about jobs should not be based on gender or race) for example, or that similarly, predictions about objects in an image should not be based on contextual information (gender of a subject in an image should not be based on the background). %Husky dogs and wolves should be discriminated regardless of background clues like snow (cite).

In this direction, we demonstrate how the Domain-Adversarial Neural Network (DANN)  developed in the context of domain adaptation \cite{ganin2016domain} can be modified to generate `agnostic' feature representations that do not incorporate any implicit contextual (correlated) information that we do not want, and is therefore unbiased and fair. {\color{black}We note that this is a far stronger requirement than simply removing protected features from the input that might otherwise implicitly remain in the model due to unforeseen correlations with other features.}

We present a series of experiments, showing how the relevant pixels used to make a decision move from the contextual information to the relevant parts of the image. This addresses the problem of relying on contextual information, exemplified by the Husky/Wolf problem in \cite{ribeiro2016should}, but more importantly shows a way to de-bias classifiers in the feature engineering step, allowing it to be applied generally for different models, whether that is word embeddings, support vector machines, or deep networks etc.

Ultimately, this ties into the current debate about how to build trust in these tools, whether this is about their predictive performance, their being right for the right reason, their being fair, or their decisions being explainable.

\section{Agnostic Models}
Methods have previously been proposed to remove biases, based on various principles, one of which is distribution matching \cite{zhao2017men}: ensuring that the ratio between protected attributes is the same in the training instances and in the testing instances. However, this does not avoid using the wrong reasons in assessing an input  but simply enforces a post-hoc rescaling of scores, to ensure that the outcome matches  the desired statistical requirements of fairness.

In our case, we do not want to have an output distribution that only looks as if it has been done without using protected concepts. We actually want a model that cannot even represent them within its internal representations, where we call such a model {\em agnostic}. This is a model that does not represent a forbidden concept internally, and therefore cannot use it even indirectly. Of course this kind of constraint is likely to lead to lower accuracy. However, we should keep in mind that this reduction in accuracy is a direct result of no longer using contextual clues and correlations that we explicitly wish to prevent.

In this direction, we consider classification tasks where $X$ is the input space and $Y = \{0,1,\dots,L-1\}$ is the set of $L$ possible labels. An agnostic model (or feature representation) $G_f: X \rightarrow \mathbb{R}^D$, parameterized by $\theta_f$, maps a data example $(\textbf{x}_i,\textbf{y}_i)$ into a new D-dimensional feature representation $\textbf{z} \in \mathbb{R}^D$ such that for a given label $p \in Y$, there does not exist an algorithm $G_y : \mathbb{R}^D \rightarrow [0,1]^L$ which can predict $p$ with better than random performance.

%This essentially works by forcing two parts of a network to compete in an adversarial manner. Think of this analogy: you want to slice a cake into exactly equal parts, but both Alice and Bob want the largest slice they can get. You can make them compete: Alice will be in charge of slicing, and Bob of assigning slices to people. This tension will ensures that we obtain an unbiased outcome in the presence of selfish behaviour. 

%The first part of the neural network can be asked to generate internal representations, which are both discriminative for the target concept, and also completely unhelpful in discriminating between two protected attributes (it could be two types of background, but could also be irrelevant properties that should not be used in the decision). The resulting network should be agnostic with respect to those attributes. This is a stronger requirement than just matching the distributional properties: not being used in making a decision is not the same as not being reflected in the decision output distribution. 

\section{Domain-Adversarial Neural Networks}
One possible way to learn an agnostic model is to use a DANN \cite{ganin2016domain}, recently proposed for domain adaptation, which explicitly implements the idea raised in \cite{ben2007analysis} of learning a representation that is unable to distinguish between training and test domains. In our case, we wish for the model to be able to learn a representation that is agnostic to a protected concept.

DANNs are a type of Convolutional Neural Network (CNN) that can achieve an agnostic representation using three components. A feature extractor $G_f(\cdot;\theta_f)$, a label prediction output layer $G_y(\cdot;\theta_y)$ and an additional protected concept prediction layer $G_p: \mathbb{R}^D \rightarrow[0,1]$, parameterized by $\theta_p$. During training, two different losses are then computed: a target prediction loss for the $i$-th data instance $\Lagr_y^i(\theta_f,\theta_y) =  \Lagr_y(G_y(G_f(\textbf{x}_i;\theta_f);\theta_y),\textbf{y}_i)$ and a protected concept loss $\Lagr_p^i(\theta_f,\theta_p) =  \Lagr_p(G_p(G_f(\textbf{x}_i;\theta_f);\theta_p),p_i)$, where $\Lagr_y$ and $\Lagr_p$ are both given by the cross-entropy loss and $p_i$ is the label denoting the protected concept we wish to be unable to distinguish using the learnt representation.

Training the network then attempts to optimise 

\begin{multline}\label{eqn:target}
E(\theta_f,\theta_y,\theta_p) = (1-\alpha)\frac{1}{n}\sum_{i=1}^n \Lagr_y^i (\theta_f,\theta_y)\\ - \alpha\left( \frac{1}{n}\sum^n_{i=1} \Lagr_p^i(\theta_f,\theta_p) + \frac{1}{n^\prime} \sum_{i=n+1}^N \Lagr_p^i (\theta_f,\theta_p) \right),
\end{multline}
using $\alpha$ as a hyper-parameter for the trade-off between the two losses and finding the saddle point $\hat{\theta}_f, \hat{\theta}_y,\hat{\theta}_p$ such that
\begin{equation}\label{eqn:min}
(\hat{\theta}_f, \hat{\theta}_y) = \argmin_{\theta_f,\theta_y} E(\theta_f,\theta_y,\hat{\theta}_p),
\end{equation}
\begin{equation}\label{eqn:max}
\hat{\theta}_p = \argmax_{\theta_p} E(\hat{\theta}_f,\hat{\theta}_y,\theta_p).
\end{equation}

As further detailed in \cite{ganin2016domain}, introducing a {\em gradient reversal layer} (GRL) between the feature extractor $G_f$ and the protected concept classifier $G_p$ allows (\ref{eqn:target}) to be framed as a standard stochastic gradient descent (SGD) procedure as commonly implemented in most deep learning libraries.

The network can therefore be learnt using a simple stochastic gradient procedure, where updates to $\theta_f$ are made in the opposite direction of the gradient for the maximizing parameters, and in the direction of the gradient for the minimizing parameters. Stochastic estimates of the gradient are made, both for the target concept and for the protected concept, using the training set.
We can see this as the two parts of the neural network (target classifier $G_y$ and protected concept classifier $G_p$) are competing with each other for the control of the internal representation. DANN will attempt to learn a model $G_f$ that maps an example into a representation allowing the target classifier to accurately classify instances, but crippling the ability of the protected concept classifier to discriminate inputs by their label for the protected concept.

\section{Experiments}
To test the use of DANNs for learning representations that can be used to make predictions for the right reasons, we ran two different experiments. In Experiment 1, we first demonstrate the issue of using contextual information to make predictions in a cross-domain classification task, before using a DANN in Experiment 2, showing that the network can learn an agnostic representation that allows us to make predictions on a target concept without using information from a correlated contextual concept, such as the image background.

\begin{figure}[t]
\centering
\includegraphics[width=0.24\columnwidth]{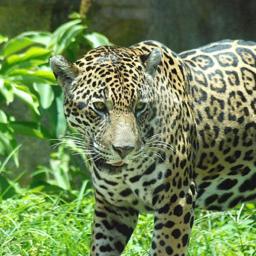}~
\includegraphics[width=0.24\columnwidth]{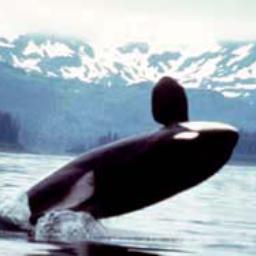}~
\includegraphics[width=0.24\columnwidth]{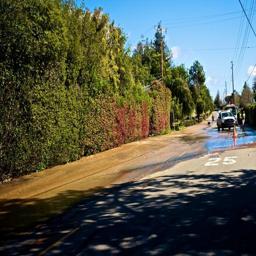}~
\includegraphics[width=0.24\columnwidth]{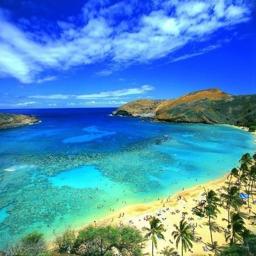}\\
\includegraphics[width=0.24\columnwidth]{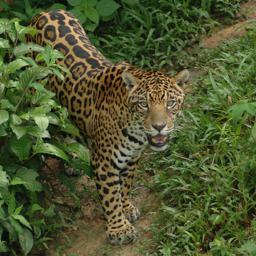}~
\includegraphics[width=0.24\columnwidth]{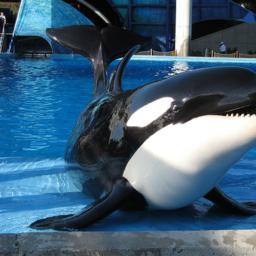}~
\includegraphics[width=0.24\columnwidth]{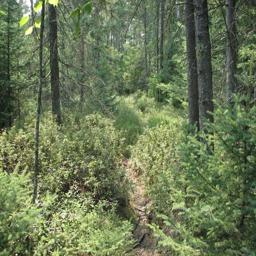}~
\includegraphics[width=0.24\columnwidth]{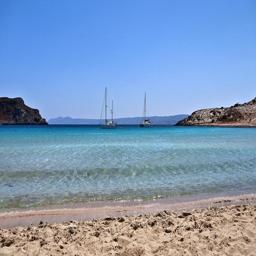}
\caption{Example images taken from the `Jaguar', `Killer whale', `Forest path' and `Coast' categories of the ImageNet and Places datasets respectively (left-right).}
\label{fig:example}
\end{figure}

\subsection{Data Description}
In this work, we combine two datasets, making use of the `Jaguar' and `Killer whale' categories from the ImageNet dataset \cite{ImageNet}, as well as the `Forest path' and `Coast' categories from the Places dataset \cite{PLACE}.

A two-part training set was constructed containing $2{,}524$ images from the `Jaguar' category, and the same number for the `Killer whale' category from ImageNet (the target concept training set). This was further supplemented with $5{,}000$ images from each of the two categories (`Forest path' and `Coast') from the Places dataset (the contextual concept training set), for a total of $15{,}048$ images in the combined training set. Two separate hold-out sets were also created, one for the target concept containing $50$ hold-out images from each of the `Jaguar' and `Killer whale' categories, and one for the contextual concept containing $50$ hold-out images from each of the `Forest path' and `Coast' categories.

Data augmentation was performed on the training set to increase the number of instances by creating new images that are multi-crops of $224 \times 224$ pixels and horizontally flipping copies of the training set images. All images in our experiments were also pre-processed to be $256 \times 256$ pixels by a process of multi-cropping where each image is resized before cropping the final size from the centre region, as in \cite{Resnet,Alexnet}. Example images from the training set used for the experiments can be seen in Fig.~\ref{fig:example}.
%{\color{blue}The selection of $\alpha$ is chosen based on empirical result on the place validation set. So we report the result of a constrained model on a separate place test set, which has the same size as the validation set.}

% \begin{figure*}[th]
% \centering
% 	\begin{subfigure}[t]{0.48\textwidth}
%     \centering
% 	\includegraphics[width=\columnwidth, height=0.75\columnwidth]{imgs/animal_alpha}
% 	\caption{}
%     \end{subfigure}
%     ~
%     \begin{subfigure}[t]{0.48\textwidth}
%     \centering
% 	\includegraphics[width=\columnwidth, height=0.75\columnwidth]{imgs/animal_acc}
% 	\caption{}
%     \end{subfigure}
%     %\\
%     %
% \caption{\label{fig:phase2}{\color{red}Accuracy of the target classifiers on the two 		validation sets for different values of $\alpha$.}}
%\end{figure*}
\begin{figure*}[t]
	\subfigure[CNN trained on the target concept training set (animals)]		{\includegraphics[width=0.48\textwidth]{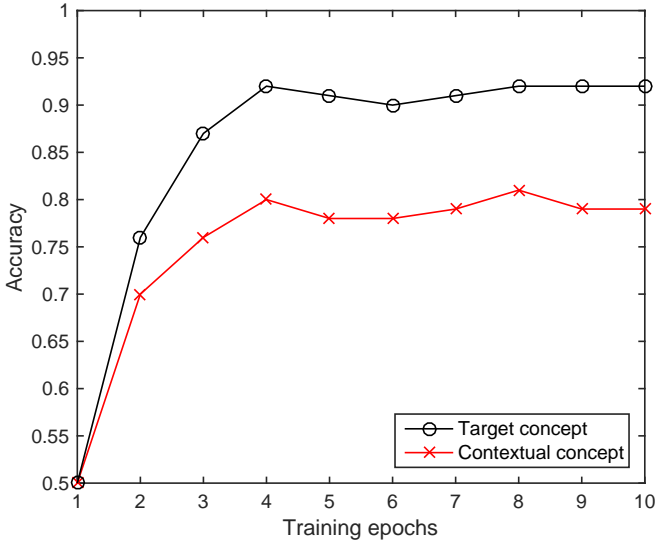}}\label{fig:1a}
    ~
    \subfigure[CNN trained on the contextual concept training set (backgrounds)]		{\includegraphics[width=0.48\textwidth]{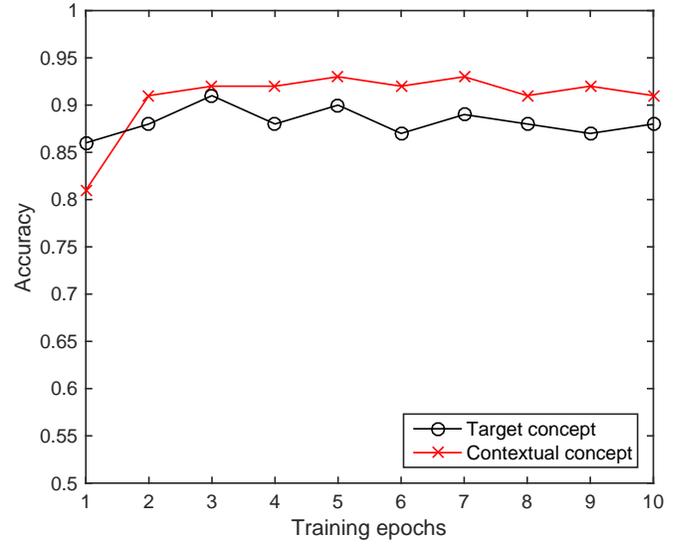}}\label{fig:1b}
\caption{\label{fig:phase1}Results from Experiment 1, showing that a standard CNN model trained on the target concept will also learn how to classify in the contextual concept and vice versa.}
\end{figure*}

\subsection{Network structure\label{sec:structure}}
The network structure used for our experiments in this paper are based upon a simplified version of the VGG-net CNN used in \cite{Vgg}, where the feature extraction layers $G_f$ consist of five convolutional layers: conv3-64\footnote{conv$a$-$b$ denotes a convolutional layer consisting of $b$ filters of size $a  \times a$.}, conv3-128, conv3-256, conv3-512 and conv3-512, with ReLU activation and max-pooling layers inserted after each convolutional layer. The output prediction classifiers $G_y$ and $G_p$ are each composed of four fully connected layers, fc-1024, with ReLU and dropout layers with a dropout of $0.5$ after each fully connected layer.

\subsection{Experiment 1: Cross-domain classification}
In this first experiment, we motivate our approach by demonstrating the problem we wish to address, namely that contextual information can be used to make classification decisions about our target concept that is not related to the target that we actually wish to learn.

We began by training from scratch two independent CNNs with the same network architecture, one on the target concept training set and one on contextual concept training set. The layers of the network are described in Sec~\ref{sec:structure} with a single output prediction classifer $G_y$ per model, $i.e.$ each CNN is composed of five convolution layers, followed by four fully connected layers with no shared features across the models. Each model was trained for $10$ epochs using the following model parameters: a batch size of $32$, a starting learning rate of $\eta=0.01$ that decays every three epochs by a factor of $10$, a momentum of $0.5$ and a weight decay of $5e^{-4}$. %{\color{blue}Note that we use all the training data to train each model, $2,524$ images for the animal model and $10,000$ for the place model.}

The accuracy of each model was measured on both the target and contextual test sets after each epoch as shown in Fig.~\ref{fig:phase1}. As one might expect, we can see that the model trained on the target concept achieves an accuracy of $92\%$ on the target test set, while the contextual concept model achieves an accuracy of $91\%$ on the contextual test set. More problematically, we can see that the target concept model, trained only on images of animals, also has good performance at classifying images of forest paths and coastlines from the contextual test set, with an accuracy of $79\%$. Similarly, the contextual concept model, trained only on images of forest paths and coastlines can correctly identify animals with an accuracy of $88\%$.

\begin{figure*}[t]
\subfigure[Performance of $G_y$ in the DANN trained for the target concept (animals)]{\includegraphics[width=0.48\textwidth]{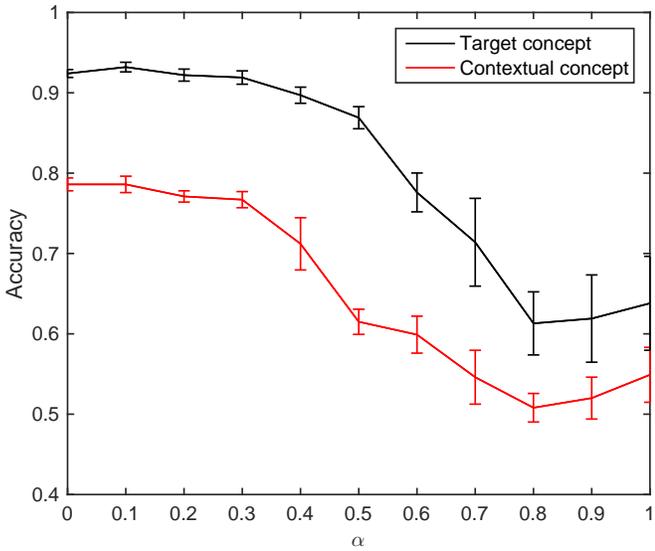}}\label{fig:2a}
    ~
    \subfigure[Performance of $G_p$ in the DANN trained for the contextual concept (backgrounds)]{\includegraphics[width=0.48\textwidth]{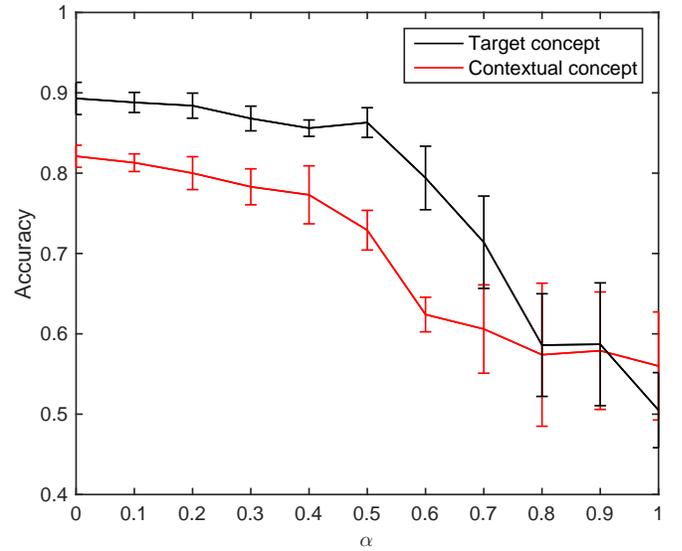}}\label{fig:2b}
\caption{\label{fig:phase2}Accuracy of the two independent classifiers in the DANN using the shared feature space on the test sets for different values of $\alpha$.}
\end{figure*}

\subsection{Experiment 2: Learning with Domain-adversarial neural networks}
In this next experiment, we show that with our proposed use of DANNs maximises its performance on the target concept whilst following the constraint that it should not learn useful features for the contextual concept. We further examine the most informative pixels (\textit{e.g.} those pixels which have the strongest response in the feature map) used for classification \cite{rcnn,Zeiler14}, showing that the most informative pixels are no longer found in the image background.

Keeping all the model parameters, apart from a new learning rate ($\eta = 0.001$), the same as in Experiment 1, we trained a single DANN model on the combined training set, with the network layers outlined in Sec~\ref{sec:structure}, with the target prediction output layers $G_y$ predicting the target concept, and the protected concept prediction layers predicting the contextual concept. By doing so, we force the model to learn a shared data representation (feature space) that maximises performance on the target while incorporating no knowledge of features which are useful for classifying the contextual concept images. This process was repeated for different gradient trade-offs in the range $\alpha = [0, 0.1, \dots, 1]$ using a grid-search procedure, where $\alpha=0$ represents simply training the share feature space on the target concept, and $\alpha = 1$ represents training the shared feature space to maximise the loss for the contextual concept.  We repeated this process $10$ times, reporting the average accuracy for each run, along with the standard deviation.

In Fig.~\ref{fig:phase2}, we can see the accuracy of the DANN for varying gradient trade-off values on the target and contextual concept test sets. Our results show that as $\alpha$ increases and is further constrained in its use of information from the contextual concept, the performance on the target concept decreases, suggesting that the performance on the intended target concept was indeed being helped by the contextual background information. Our results show that once we have removed features which are useful for predicting the contextual concept, our target classifier achieves an accuracy of $64\%$, while the contextual classifier can only maintain an accuracy close to random guessing.

\begin{figure*}[t]
\centering
\subfigure[Activation maps for the `Killer whale' category]{\includegraphics[width=0.48\textwidth]{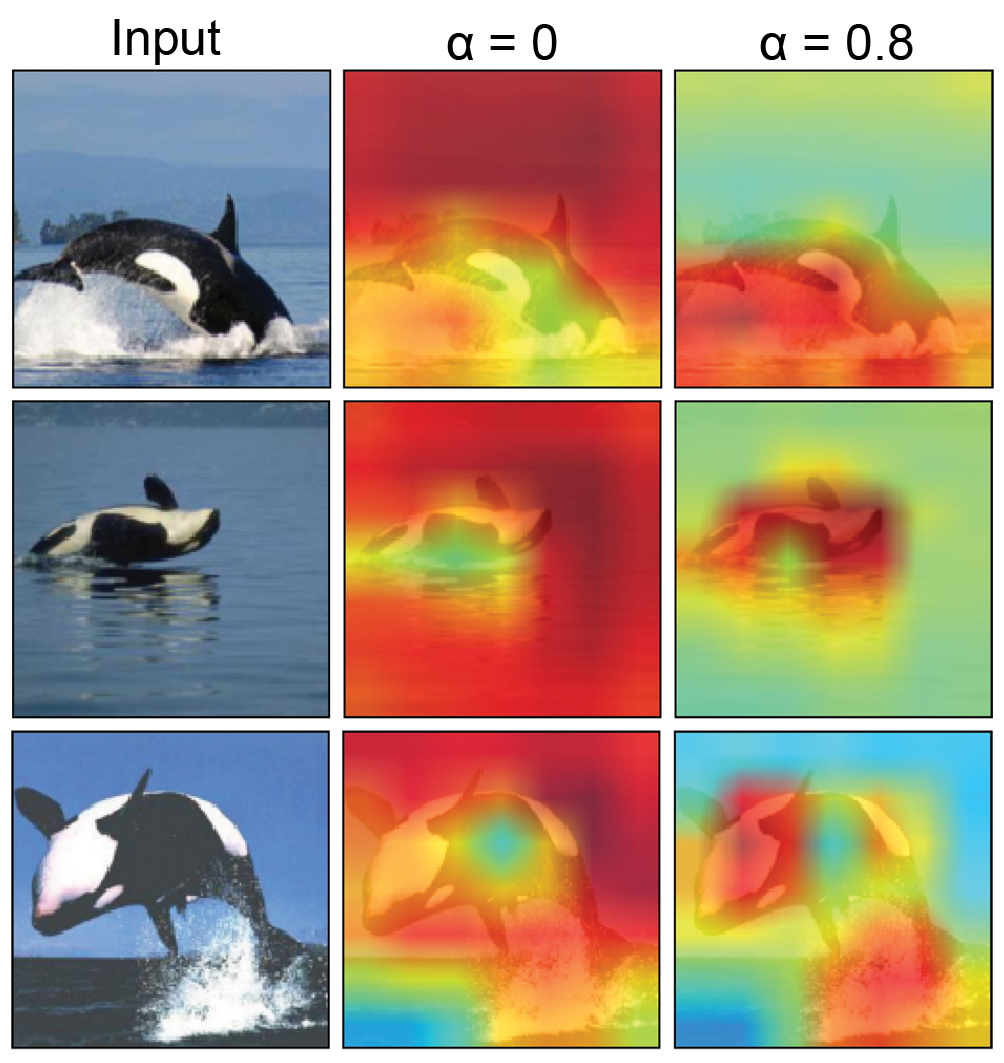}}
    ~
\subfigure[Activation maps for the `Jaguar' category]{\includegraphics[width=0.48\textwidth]{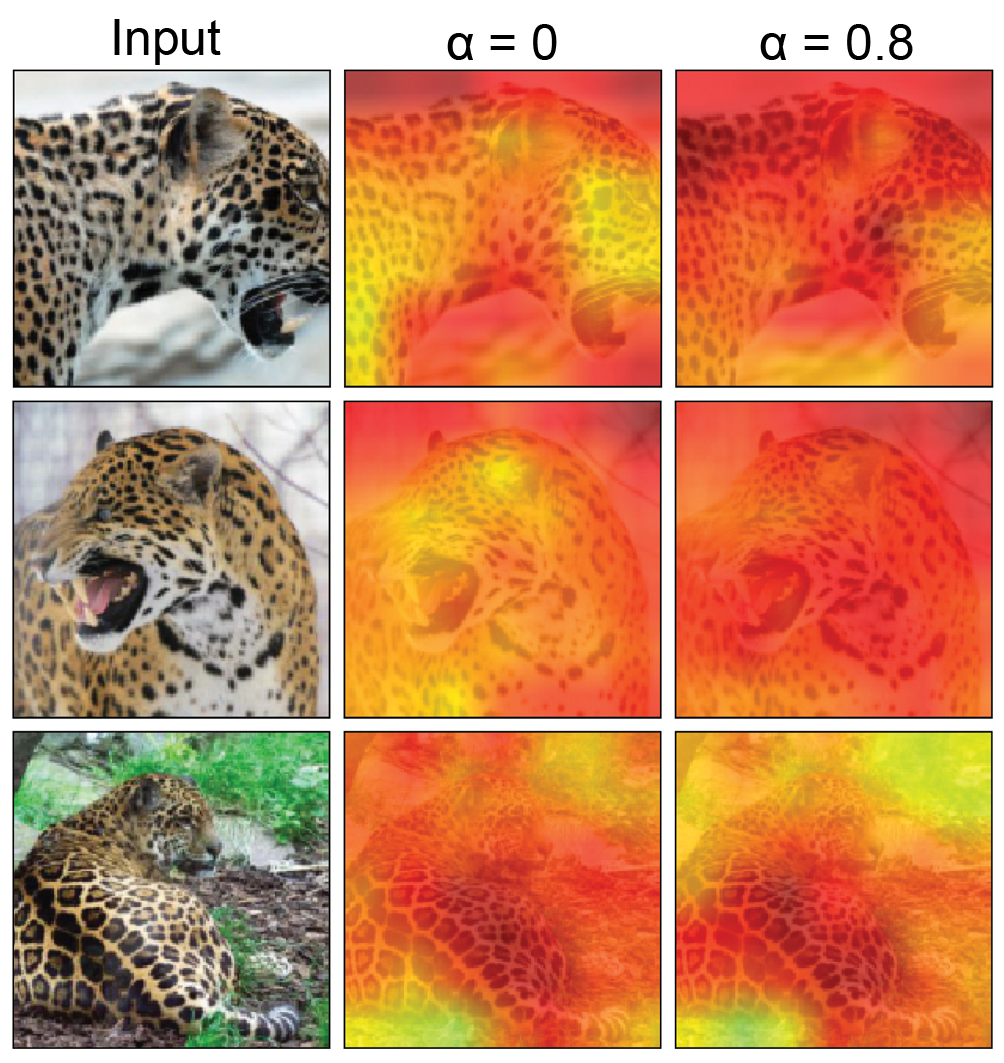}}
\caption{{\color{black}Activation maps based on the strongest response of the shared feature representation. Examples selected are those with the least correlation between the activation maps for $\alpha = 0$ and $\alpha = 0.8$ as shown in the images.}}
\label{fig:pixels}
\end{figure*}

We further investigated whether after applying the minimax procedure of the DANN that the most informative pixels for prediction corresponded with the location of the target concept in the image, or whether they were focused on the background scene of the image. Fig.~\ref{fig:pixels} shows activation maps for the feature representation shared between the independent classifiers on a set of three images for each target concept category. Examples were selected as those with the least correlation between the activation maps for the contrasting $\alpha$ values of $0$ and $0.8$ shown, where $\alpha$ values were chosen as the two extremes in the classification accuracy.

We can observe that for the `Killer whale' category, the most informative pixels for $\alpha = 0$ are indeed found in the background of the image, while for $\alpha = 0.8$ the activation maps show that the network is focusing on the actual body of the animal instead, as desired. For the `Jaguar' category, analysis of the most informative pixels is less clear, with activation generally being spread widely across the image. However, we do see some evidence of a stronger activation response to parts of the jaguar's body overall.

\section{Discussion}
In our experiments, we found that as the model learns the shared features with increasingly less contextual information, accuracy of the target classifier decreases. This {\color{black} is exactly what we expect and directly addresses our main argument, that previously the classifier was relying on the contextual background information that should not be used to make its predictions.}

At one extreme, where $\alpha = 1$, the network is using no information from the target concept in its data representation, instead trying to maximise its loss on the contextual concept in the shared feature space $G_f$, while minimising its loss in the contextual classifier $G_p$. This tension between the two parts of the network leads to a minimax scenario where if there is any information which can be exploited to correctly predict in the contextual concept, it is subsequently removed from the data representation. 

We note that ideally $\alpha$ should be set to $1$ for similar experiments, given that for any other setting the learning system would still be exploiting forbidden information, and would not be satisfying the original requirements of the task: to learn to predict without the contextual information. However, since in this scenario the shared feature space would not rely on the target domain at all, $\alpha$ needs to be slowly increased as training progresses until reaching its maximum. In this way, the features will be guided by the target domain as well, forming a saddle point in the exploration of the feature space as required.

{\color{black} Results from investigating the most informative pixels for classification at differing levels of $\alpha$ revealed that the constraint of the contextual concept appears to have been more successful for the `Killer whale' and `Coast' images than for the `Jaguar' and `Forest path' pairing. This can perhaps be best explained by how closely the contextual concept training images represent the contextual concept found in the target concept training images, \textit{i.e.} the whales are always pictured next to or in the ocean, whereas jaguars will sometimes be found outside of the jungle with different backgrounds, and therefore the `Forest path' category does not match `Jaguar' backgrounds as closely as `Coast' does for the `Killer whale' category.}

Further theoretical and experimental analysis of additional minimax architectures is needed to explain the phenomena of the target classifier accuracy increasing on both target and contextual test sets for values of $\alpha \geq 0.8$.

\section{Conclusions}
{\color{black} The creation of a new generation of AI systems that can be trusted to make fair and unbiased decisions is an urgent task for researchers. As AI rapidly conquers technical challenges related to predictive performance, we are discovering a new dimension to the design of such systems that must be addressed: the fairness and trust in the system's decisions.}

{\color{black} In this paper, we address this critical issue of trust in AI by not only proposing a new high standard for models to meet, being agnostic to a protected concept, but also proposing a method to achieve such models.} We define a model to be agnostic with respect to a set of concepts if we can show that it makes its decisions without ever using these concepts. This is a much stronger requirement than in distributional matching or other definitions of fairness. We focus on the case where a small set of contextual concepts should not be used in decisions, and can be exemplified by samples of data. 
We have demonstrated how ideas developed in the context of domain adaptation can deliver agnostic representations that are important to ensure fairness and therefore trust.

Our experiments demonstrate that the DANN successfully removes unwanted contextual information, and makes decisions for the right reasons. While demonstrated here by ignoring the physical background context of an object in an image, the same approach can be used to ensure that other contextual information does not make its way into black-box classifiers deployed to make decisions about people in other domains and classification tasks.

\section*{Acknowledgements}
SJ, TLW and NC are support by the FP7 Ideas: European Research Council Grant 339365 - ThinkBIG.

\end{document}